\documentclass[ journal]{IEEEtran}

\usepackage{ifthen}
\usepackage{amsmath,bm}
\IEEEoverridecommandlockouts       

\usepackage[utf8]{inputenc}
\usepackage{graphicx}
\usepackage{amsmath}
\usepackage{algorithm}
\usepackage[noend]{algpseudocode}
\usepackage{multirow}
\usepackage{subcaption}
\usepackage{verbatim}
\usepackage{color,soul}

\usepackage{multirow}
\usepackage[dvipsnames]{xcolor}

\usepackage[switch]{lineno}
\modulolinenumbers[2]
\usepackage{cite}

\usepackage{subcaption}
\usepackage{booktabs}
\usepackage{dirtytalk}

\begin{document}
\title{Improved Soft Actor-Critic: Mixing Prioritized Off-Policy Samples with On-Policy Experience}
%
%
\author{Chayan Banerjee$^1$,
Zhiyong Chen$^1$, and Nasimul Noman$^2$
\thanks{$^1$School of Engineering, $^2$School of Information and Physical Sciences,
        University of Newcastle, Callaghan, NSW 2308, Australia. 
        Z. Chen is the corresponding author. Email: {\tt\small  zhiyong.chen@newcastle.edu.au}}%
}
%
%
%
\maketitle              
\begin{abstract}
Soft Actor-Critic (SAC) is an off-policy actor-critic  reinforcement learning algorithm, essentially based on entropy regularization. SAC trains a policy by maximizing  the  trade-off  between  expected  return and  entropy (randomness in the policy). It has achieved state-of-the-art performance on a range of continuous-control benchmark tasks, outperforming prior on-policy and off-policy methods. SAC works in an off-policy fashion where data are sampled uniformly from past experiences (stored in a buffer) using which parameters of the policy and value function networks are updated. We propose certain crucial modifications for boosting the performance of SAC and make it more sample efficient. In our proposed improved SAC, we firstly introduce a new prioritization scheme for selecting better samples from the experience replay buffer. Secondly we use a mixture of the prioritized off-policy data with the latest on-policy data for training the policy and the value function networks. 
We compare our approach with the vanilla SAC and some recent variants of SAC and show that our approach outperforms the said algorithmic benchmarks. It is comparatively more stable and sample efficient when tested on a number of continuous control tasks in MuJoCo environments. 
 \end{abstract}
 
\begin{IEEEkeywords}
Soft Actor critic, reinforcement learning,  policy optimization, on-policy learning, off-policy learning
\end{IEEEkeywords}

\section{Introduction}
Reinforcement learning (RL) has showed significant growth and achieved state-of-the-art performance in diverse domains including robotics \cite{levine2016end,neunert2020continuous}, locomotion control \cite{peng2017deeploco, xie2018feedback}, strategy games \cite{shi2018adaptive,vinyals2019grandmaster}, and so on. In the conventional model-free RL paradigm, an agent can be trained either by learning an approximator of action-value (Q) function  \cite{mnih2015human, bellemare2017distributional} or by explicitly learning and optimizing a policy model.  In particular, the class of actor-critic (AC) algorithms \cite{konda2000actor,mnih2016asynchronous}  follows the later approach. In a policy optimization process, a typical AC algorithm performs  three basic steps of  1) generating samples by running current policy on environment, 2) estimating a state/action-value function from the samples, and 3) evaluating the performance of the last action and updating the current policy. 

Based on how the past experiences are utilized the AC or the RL algorithms in general can be classified as either on-policy or off-policy algorithms. An on-policy algorithm continually learns a policy from the samples generated by the policy upon interacting with the environment but discards them once used. Popular examples of on-policy algorithms include vanilla policy gradient \cite{sutton1999policy}, trust region policy optimization\cite{schulman2015trust}, proximal policy optimization \cite{schulman2017proximal}, and so on. By contrast, an off-policy algorithm learns a policy from the samples of past experiences stored in a buffer, e.g., deep deterministic policy gradient (DDPG) \cite{silver2014deterministic,lillicrap2015continuous}, twin delayed DDPG \cite{fujimoto2018addressing} and soft actor-critic (SAC) \cite{haarnoja2018soft}. In particular, SAC is a model-free off-policy algorithm that optimizes a stochastic policy, with the use of entropy regularization.  Increase in the policy's entropy results in more exploration and accelerates learning. SAC has so far achieved state-of-the-art performance on a number of continuous control benchmark tasks, outperforming several on-policy and off-policy algorithms. The experimental results in \cite{haarnoja2018soft} 
demonstrate that SAC outperforms the benchmark algorithms in the aforementioned references \cite{schulman2017proximal,silver2014deterministic,lillicrap2015continuous,fujimoto2018addressing} and soft Q-learning in \cite{schulman2017equivalence}.

As an  off-policy algorithm, SAC maintains a so-called experience replay (ER) buffer and trains its policy network
on the data uniformly sampled from the buffer.  
ER forms the basis of most off-policy algorithms where  actors and/or critics are updated according to the samples 
from the past experiences stored in an ER buffer.  Using ER greatly increases the sample efficiency of an algorithm by enabling data to be reused multiple times for training the policy.  
It typically needs a large number of environment-agent interactions to obtain experiences (or transition tuples) and maintain the ER buffer. Researchers have studied different approaches of leveraging the ER buffer for policy optimization in literature. 

For instance, a sample efficient AC algorithm with ER \cite{wang2016sample} interleaves the on-policy learning and off-policy learning cycle where a hyper-parameter controls the ratio of off-policy updates to on-policy updates. 
Hindsight  ER \cite{andrychowicz2017hindsight} and its recent variants \cite{fang2019curriculum,fang2018dher,nguyen2019hindsight} aim to essentially deal with sparse reward and multi-goal RL environments. 
The remember and forget for ER \cite{novati2019remember} algorithm enforces the similarity between policy and the experiences in the replay buffer. It characterizes past experiences as either \say{near-policy} or \say{far-policy} based on the deviation from the importance sampling weight and calculate gradients only from \say{near-policy} experiences. In a different work, Schimit et. al.\cite{schmitt2020off} introduces a Q-function free AC paradigm where
concurrent agents share their experiences through a common ER module and also use V-trace importance sampling \cite{espeholt2018impala} for variance reduction. The research shows that mixing of a proportion of on-policy experiences with off-policy experiences contributes to the improvement of the proposed algorithm's convergence.

%

It is intuitively known that  differentiating important samples from not so important ones is beneficial to policy learning; see, e.g.,  \cite{katharopoulos2018not}. Prioritized experience replay (PER) is a strategy for differentiating samples in ER.
In \cite{schaul2015prioritized}, the PER algorithm samples non-uniformly from the replay buffer and favors those samples which have a higher value of absolute temporal difference (TD) error. 
PER has also been tested with double deep Q-networks (DQNs) \cite{van2016deep}, dueling network architecture DQNs \cite{wang2016dueling}, and DQNs with  snapshot ensembling \cite{schulze2018vizdoom}, and it performs better than the non-prioritized/ uniform sampling approach. As a result, PER has been adopted in many DQN extensions,  e.g., \cite{hessel2018rainbow, horgan2018distributed, daley2019reconciling, pan2018organizing}. In a recent DQN-PER extension, called prioritized sequence experience replay \cite{brittain2019prioritized}, the algorithm not only assigns high sampling priority to important transitions but also increases the priorities of the previous transitions that lead to important transitions. 
PER has also been applied to the AC type of algorithms, e.g., for DDPG \cite{8122622}.

Apart from using absolute TD error for sample prioritization, episodic return / episodic memory has also been used for the same purpose in a number of works \cite{lin2018episodic, lee2019sample, zhang2019asynchronous}. For example, the asynchronous episodic DDPG \cite{zhang2019asynchronous} improves over the vanilla DDPG algorithm by using two replay buffers and sampling from them with a fixed probability for training the DDPG networks. One buffer (the Memory) is just like the generic DQN style replay buffer and other (the HMemory) stores only those episodic experiences whose episodic reward (or score) surpasses the best score at present. The work highlights the demerits of learning the policy with the episodic experiences from episodes with a \say{best so far} score. This approach leads to reduction in sample diversity which might result in over-fitting of the learnt networks and introduction of bias for the final convergence.

In this paper, we aim to propose a strategy to manage an ER buffer as enhancement to 
the vanilla SAC algorithm to reduce the number of these environment-agent interactions and thus increase the speed of learning. 
The new version of SAC is named improved SAC (ISAC) throughout the paper, which has two primary innovations. On one hand, we introduce a  priority based selection of better quality data from mini-batches sampled from the ER buffer. On the other hand, we strategically mix the latest on-policy experiences with the current prioritized mini-batch of samples and then train the SAC networks with the updated batch. We compare our approach with the vanilla SAC and other latest SAC versions on several MuJoco continuous control tasks and our approach shows superior performance in comparison.

The remaining sections of the paper are organized as follows. In Section \ref{prelims} we discuss the preliminaries and motivations of our work. In Section \ref{details} we discuss in detail our proposed ISAC algorithm. Section \ref{Experiments} contains the experiments on several MuJoco continuous control tasks with some discussion about the results with regards to compared benchmarks. Finally in Section~\ref{concl} we conclude this paper and discuss some future extensions of the work.

\section{Preliminaries and motivation}\label{prelims}
The paper is concerned about a Markovian dynamical system represented by a
conditional probability density function $p(s_{t+1}|s_t,a_t)$
where  $s_t  \in \mathcal{S}$ and $a_t \in \mathcal{A}$ 
are the current state and action respectively at time instant $t = 1,2,\dots$, 
and  $s_{t+1} \in \mathcal{S}$ represents the next state at $t+1$.
Here, $\mathcal{S}$ and $\mathcal{A}$ represent the continuous state and action spaces, respectively.
The objective is to learn a stochastic policy $\pi_{\phi}(a_t|s_t)$ parameterized by $\phi$.
Now, the closed-loop trajectory distribution for the episode $t=1,\dots, T$ can be represented by 
\begin{align*}
   p_{\phi}(\tau) = &{p_{\phi}(s_1,a_1,s_2,a_2,\dots,s_T, a_{T},s_{T+1})} \\
  = & p(s_1)\prod_{t=1}^T 
     \pi_{\phi}(a_t|s_t) p(s_{t+1}|s_t,a_t) .
\end{align*}
Denote  $r_t = R(a_t, s_{t+1})$ as the reward generated at time $t$. 
The objective is to find an optimal policy, represented by the parameter
 \begin{align*}
     \phi^* = \arg\max_{\phi}\, \underbrace{{\mathbf E}_{\tau \sim p_{\phi}(\tau)}\, \Big[\Sigma_{t=1}^T\,\gamma ^{t} R( a_t,s_{t+1})\Big]}_{J(\phi)},
     \end{align*}
which maximizes the objective function $J(\phi)$, $\gamma$ is the discount factor. 
 
 SAC uses a maximum entropy objective, formed by augmenting the typical RL objective with the expected entropy of the policy over $p_{\phi}(\tau)$. In other words, an agent receives an additional reward at each time step which is proportional to the policy's entropy at that time-step, given by   $\mathcal{H}(\pi_{\phi}(\cdot|s_{t}))$. So, the SAC's entropy-regularized RL objective to find an optimal policy can be written as:
\begin{align*}
    \phi^{*} = \arg\max_{\phi} \mathbf{E}_{\tau \sim p_{\phi}(\tau)} \big[ \Sigma_{t}^{T} \gamma ^{t}\big(R(a_t, s_{t+1}) + \alpha  \mathcal{H}(\pi_\phi(\cdot|s_{t}))\big)\big],
\end{align*}
where the entropy regularization coefficient  $\alpha$ determines the relative importance of the entropy term against the reward.
Throughout the research in this paper, we use the latest version of SAC \cite{haarnoja2018soft2} where $\alpha$ varies over the course of training and is not fixed. 

Following the generic AC framework, SAC learns a policy $\pi_{\phi}$ which takes in the current state and generates the mean and standard deviation of an action distribution (defining a Gaussian). But instead of a single Q-network, the SAC concurrently learns two Q-networks $ Q_{\psi_1}, Q_{\psi_2}$ by regressing to the values generated by a shared pair of target networks $Q_{\psi_{targ,1}}, \; Q_{\psi_{targ,2}}$ \cite{haarnoja2018soft2}.
%
SAC works on an off-policy scheme and thus uses an ER buffer to update network parameters. It alternates between a \say{data collection} phase and a \say{network parameter update} phase. In the data collection phase, SAC saves to ER buffer the transition tuples, e.g. $d_t^e = (s_t,a_t,r_t,s_{t+1}),\; \text{where}\; t = 1,\dots,T_e, \; e= 1,\dots,E$ ($E$ is total episodes run),  obtained by running the current policy in the environment. In the network update phase, SAC samples a mini-batch ($B$) of saved transition tuples from the ER buffer ($B \sim \mathcal{D}$) uniformly and updates the network parameters. For more details on SAC, please refer to \cite{haarnoja2018soft,haarnoja2018learning,haarnoja2018soft2}.

\begin{figure}[h]
\centering
\includegraphics[width=0.4\textwidth]{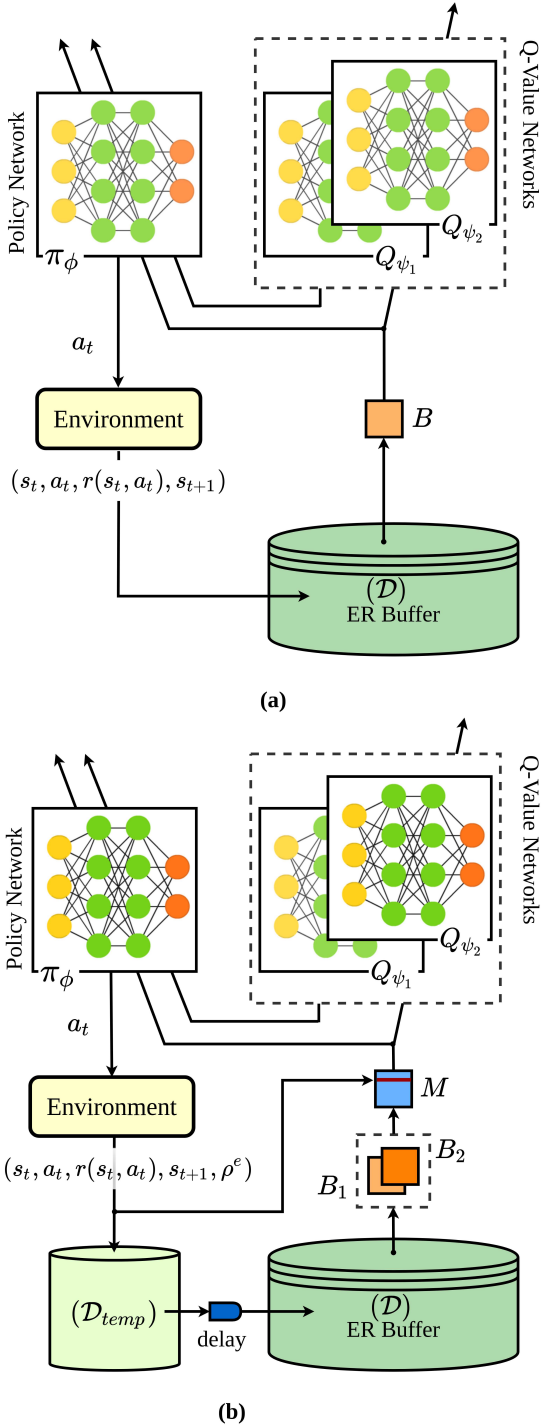}
\caption{Simplified illustrations of (a) the vanilla SAC \cite{haarnoja2018soft2} and (b) proposed improved SAC (ISAC). The target Q network(s) are not shown in the diagram for clarity.}  \label{SAC block diagram}
\end{figure} 

On the subject of improving SAC, there has been a number of recent works, where the authors have targeted different aspects of the SAC framework. E.g. Ward et. al. \cite{ward2019improving} altered the  choice of policy distribution from factored Gaussian in vanilla SAC to Normalizing flow policies for improving exploration. Campo et. al. \cite{campo2020band} explicitly constrained the ability of the critic to learn the high frequency components of the state-action value function through the addition of a convolutional filter. 

On the contrary to these works, we target the problem of utilizing the ER buffer more efficiently such that the SAC learns a well generalized policy, fast and thus with lesser agent-environment  interactions (or samples). In this regard, \cite{wang2019boosting} and \cite{sinha2020experience} have experimented with introducing TD error based prioritization (i.e. PER \cite{schaul2015prioritized}) in vanilla SAC and achieved  improved performance.  The authors of \cite{wang2019boosting} proposed an emphasizing recent experience (ERE) approach, where the algorithm samples more aggressively from recent experiences such that the updates from old experiences do not overwrite those from new experiences. 
The combination of the aforementioned algorithms, i.e., SAC+PER+ERE  shows superior performance in a number of continuous control tasks. In a slightly different approach proposed by Sinha et.al. \cite{sinha2020experience}, the algorithm works on the lines of \cite{zhang2019asynchronous}, by maintaining two replay buffers. Here, one  (fast) buffer stores the recent-most on-policy experiences and another (slow) buffer contains additional off-policy experiences. Next, they estimate density ratios between  near off-policy and near on-policy distribution, by minimizing an objective over a density estimator network. They use this ratio as weight over the Q-network update objective so as to encourage more updates over samples closer to the fast replay buffer. Though it shows superior performance in certain benchmark environments, it introduces a number of additional hyperparameters including the size of the fast replay buffer and the architecture of the density estimator network. Besides the density ratio estimation process also adds computational complexity to the vanilla SAC.

In this paper, we propose an approach to make ER buffer based training of SAC networks more sample efficient, with increased stability in policy performance and faster convergence to near-optimal performance. But instead of prioritized sampling of recent data like in \cite{wang2019boosting} or calculating importance weights using multiple buffers and additional networks like \cite{sinha2020experience}, we introduce a different and simpler approach as elaborated below.

Firstly, we introduce a prioritization scheme to improve the quality of off-policy samples  from an ER buffer. By using an episodic return based priority score to select better transitions, we obtain a prioritized off-policy batch for network training. Regarding the use of episodic return to prioritize ER buffer samples and checking overfitting of networks while training frequently on better performing samples, we draw a number of crucial insights from these works \cite{8122622,lin2018episodic, lee2019sample, zhang2019asynchronous}.  Secondly we mix the latest on-policy experiences directly into this prioritized off-policy batch and train the SAC networks (at each timestep). 
An itemized list of out major contributions to the existing conventional SAC framework is summarized below.

\begin{enumerate}
    \item  Sampled data prioritization (SDP): We save all the  transition  tuples (to the ER buffer $\mathcal{D}$) with an additional element. It is the episodic return  $\rho$ (cumulative reward of an episode) of a certain transition tuple's parent episode and serves as a goodness or priority score for a transition. At each timestep we accumulate data or transitions from multiple mini-batches uniformly drawn from the ER buffer. From the accumulated data we choose a prioritized batch based on the priority score. We also control prioritized selection using a semi-metric calculation of $\rho$. 
    
    \item Mixing on/off policy experiences (MO/O): The pure off-policy with sampled and prioritized training data is then mixed with the latest on-policy transition. This mixed batch is used to train the policy and the value function networks of SAC.
    
    \item Delayed infusion of recent experience: We add the recent most collection of experiences/ transitions to the ER buffer in a delayed fashion. This enables the SDP to calculate and assign the priority score. This also helps preserve data diversity in the ER buffer, especially for the training on mixture of on-off policy experiences.
    
    \end{enumerate}
    
To the best of our knowledge, this is the first paper that considers direct on- and off-policy data mixing in the context of SAC algorithm. It is also the first paper that  considers prioritizing data that have been uniformly sampled from the ER buffer instead of going for a prioritized sampling approach in the first place.

\section{Improved Soft Actor Critic: SAC+SDP+MO/O} \label{details}

The proposed approach is called improved SAC (ISAC) that involves mixing the latest on-policy data with a prioritized batch of off-policy data (SDP and MO/O) and training the SAC networks with it, also denoted as SAC+SDP+MO/O. 
The approach also includes  delay in infusion of recent experiences to the ER buffer in order to improve the stability and for maintaining better data diversity in the ER buffer. In Fig. \ref{SAC block diagram} we illustrate the overall structure of the proposed methodology. 
We elaborate these features in the following subsections.

\subsection{Sampled data prioritization (SDP)} \label{SDP}

Firstly, we introduce an additional element to the experience/transition tuple added to the ER buffer in conventional SAC. This element is later used for prioritizing samples. We use the episodic return $(\rho^e)$ of a terminated episode with index $e$, as a score of goodness of all the individual transitions $(d_1^{e},\dots,d_t^{e})$ under that certain episode, 
where $d = {(s_t,a_t,r(a_t,s_t),s_{t+1})}$. In the augmented tuple $\hat{d}= {(s_t,a_t,r(a_t,s_t),s_{t+1}, \rho})$, the score $\rho^e$ remains same for all the transitions of a certain episode $e$. 
It is worth noting that unlike conventional prioritization approaches, e.g. PER, we do not sample data from the ER buffer based on any priority. We rather do prioritize data selection from some collected data which are first sampled uniformly from the ER buffer.
We uniformly sample multiple mini-batches (with replacement) and then prioritize them based on the data in those mini-batches.
More specifically, we extract $l (=2)$ of $k$ sized mini-batches from the ER buffer, i.e.,  $B_1 = [\hat{d}_1^1,\dots,\hat{d}_{k}^1], \, B_2 = [\hat{d}_1^2,...,\hat{d}_{k}^2]$. Next, we merge the batches, i.e. $\hat{C} = B_1 \cup B_2$, resulting in $k_{eff} = l \times k$ transition tuples. Then,  we pick the prioritized data $C_{prior}$ of size $k$, with $ C_{prior}\subset \hat{C} $ and $C_{rest} = \hat{C}\backslash C_{prior}$, such that $\rho_i^{e_{i}} > \rho_j^{e_{j}}$ for all $\hat{d}^{e_i}_i\, \in C_{prior}$ and $\hat{d}^{e_j}_j\, \in C_{rest}$. 
 Selecting prioritized data from a comparatively larger $\hat{C}$ with a large $l$ 
essentially trains the networks with a higher percentage of repeated data. So using $l=2$ gives a superior data diversity, while keeping the mini-batch size and other parameters unchanged.

We also introduce mechanisms to control prioritization since learning only from better performing experiences or transition tuples may cause overfitting of the networks, as noted in, e.g., \cite{zhang2019asynchronous}. Moreover, the episodic return based score values ($\rho^e$) of the transition tuples improve over time with the policy and successive episodic return values can become very similar or same in certain cases. So, as time goes the ER buffer is filled with more transition tuples with similar score values. In this situation,  if we sample minibatches with similar score values then performing prioritization on the data becomes ineffective.  Thus, non-prioritized samples should help  train the networks better. Therefore,  if the two sampled mini-batches $B_1$ and $B_2$ are similar with respect to their scores then we do not perform prioritization but use uniformly sampled mini-batch for training in that timestep.  

To be precise, we do the prioritization step (SDP) and obtain a prioritized off-policy  batch, i.e. $C = C_{prior}$, if the condition ${\zeta \leq \zeta_{th}}$ is satisfied, where $\zeta$ is a calculated similarity value between the two mini-batches and $\zeta_{th}$ is a preset threshold hyperparameter. Otherwise we choose any one batch randomly from the sampled mini-batches $B_1, B_2$, i.e.,
\begin{align*}
C = \left\{ \begin{array}{ll}  C_{prior}, & \zeta \leq \zeta_{th} \\
B  \in \{B_1,\, B_2 \}, & \zeta > \zeta_{th}
\end{array} \right. .
\end{align*}
We calculate a cosine similarity measure $(\zeta)$ of the sampled mini-batches $ (B_1\, , B_2)$ using their respective episodic return score vectors, i.e., $v_{B_{1}} = \{\rho_1,\dots,\rho_{k}\}_1$ and $v_{B_{2}} = \{\rho_1,\dots,\rho_{k}\}_2$. In particular, the cosine similarity is given by the following expression:
\begin{align*}
    \zeta = \cos ^{-1}\bigg( \frac{v_{B_{1}} \cdot v_{B_{2}}}{\|v_{B_{1}}\|\, \| v_{B_{2}}\|}\bigg)
\end{align*}
where $\|v_{B_1}\| $ and $\| v_{B_2}\| $ are the Euclidean norms of the corresponding vectors. The closer the $\zeta$ value to $1$, the greater the match between the vectors and the more similar the batches are to one another.  
Setting a higher threshold value $\zeta_{th}$ allows more prioritization of sampled data, leading the networks to learn on a higher percentage of better performing transition tuples. We examined the effect of varying the value of $\zeta_{th}$ on the performance of ISAC and the results are presented in Section~\ref{Experiments}.


\subsection{Mixing on/off policy experiences (MO/O)} \label{ME}

During the parameter update phase, the conventional SAC trains its networks by sampling one mini-batch per gradient step, from the ER buffer. This mini-batch thus contains samples of experiences collected over past iterations of the policy. We make changes to the sampled data by including the latest on-policy experiences  into them and train the networks with the revised data. 
In our approach, after the prioritization phase (SDP) we obtain a batch $C$ and then include the latest on-policy experiences to it. We replace a random tuple $y\in C$ with the latest on-policy transition tuple $q_t$, which gives us the final batch to train the SAC networks for the current timestep, i.e., $M = (C\backslash\{y\})\cup\{q_t\} $.  

MO/O improves the performance of the policy by ensuring   that the networks are trained using the latest on-policy samples mixed with the prioritized off-policy batch. There is another crucial contribution of MO/O. As mentioned in Section~\ref{SDP} with the goodness score $\rho$, we need to wait for at least one episode to generate the score, add the score to the relevant transition tuples, and then save the augmented transition tuples to the ER buffer. During the episode length no new latest transitions or experiences are added to the ER buffer and so the learning process will continue on the samples drawn from a collection of relatively older experiences. But inclusion of the latest experiences in the off-policy samples on the go resolves this issue. We have studied the performance of SAC without the MO/O type data mixing and the empirical results show slow policy learning without the MO/O artifact. The detailed results are discussed   in Section~\ref{Experiments}.

\subsection{Delayed infusion of recent experiences }\label{delay}

The SDP scheme needs to generate episodic return based scores $\rho$ for use in prioritization. 
Assigning $\rho$ to the transitions collected over an episode is based on the cumulative rewards (episodic return) accumulated over that episode. So at  least we need to wait through one episode to generate and assign the $\rho$ score.
To enable this process we employ a delayed infusion of recent experiences strategy. I.e. we add the recent on-policy experiences to the ER buffer in a delayed fashion contrary to the conventional practice of implementing it per timestep. 
The other reason for delayed infusion of data to the ER buffer is to retain its data diversity.
Using MO/O, we train the networks with mix of the latest on-policy experiences and the sampled past experiences. 
If we simultaneously add the same latest experiences to the buffer then we increase the probability of the networks seeing more of same data, especially in the beginning when the ER buffer is relatively empty. So delaying the data addition process aids in building and retaining the data diversity of the off-policy samples drawn from the ER buffer.



The detailed procedure of delayed infusion of recent experiences is elaborated as follows.
During the data collection phase of the conventional SAC framework, one on-policy experience tuple (latest interaction with the environment) is added to the ER buffer at each timestep. We delay this  addition of the latest experiences to the ER buffer.
Specifically, we save the recent experiences from the environment into a temporary buffer $\mathcal{D}_{temp}$ for $\xi$ episodes,
i.e., $\mathcal{D}_{temp} \leftarrow \{(\hat{d}_1^{1},\dots,\hat{d}_t^{1}), \dots, (\hat{d}_1^{\xi},\dots,\hat{d}_t^{\xi}))\}$, and  then push the whole data accumulated over the last $\xi$ episodes to the ER buffer, i.e., $\mathcal{D} \leftarrow \mathcal{D}_{temp}$, for every $\xi$ episodes.  The temporary buffer is reset after every transfer. 

%
Regarding the choice of the parameter $\xi$, we consider two extreme cases.  Suppose $\xi=1$, then as explained above frequently adding past (one episode worth) on-policy data to the buffer while running the MO/O style training will lead to the learned policy being overfitted. On the contrary, if $\xi$ is very high, e.g. $>20$, then learning of the network becomes very slow since we do not add new data to the ER buffer for a longer time and the networks learn mostly from the samples of very old experiences. 
We pick a reasonable value of $\xi=10$ for the experiments discussed in in the next section. 

\section{MuJoco Experiments} \label{Experiments}

We run experiments on three MuJoco \cite{6386109} continuous control tasks, InvertedPendulum-v2, Reacher-v2, 
and Swimmer-v2,  implemented in OpenAI Gym \cite{brockman2016openai}. The objective is to 
compare our proposed ISAC (SAC+SDP+MO/O) algorithm with three SAC variants, namely the vanilla SAC \cite{haarnoja2018soft2}, SAC+PER  \cite{schaul2015prioritized}, and SAC+ PER+ ERE \cite{wang2019boosting}.  
In particular, we aim to verify that  ISAC can attain early convergence to near-optimal performance
and ISAC requires substantially less number of environment-agent interactions or training samples to reach high performance.

%

We use the same neural network architecture, hyperparameters, replay buffer size, etc., in all the algorithms for fair comparison.
The important design parameters are provided in Table \ref{table:parameter}.
The code repository \cite{SAC} is used for the implementation of all the algorithms.
After a policy is trained for every certain steps (called one training unit for convenience), its performance is immediately evaluated by running the corresponding deterministic policy (i.e. the mean policy) for five consecutive episodes. 
One training unit is $1,000$ steps for InvertedPendulum-v2 and Swimmer-v2, and it is $500$ steps for Reacher-v2.
Totally, we run $N=100,400,500$ training units (the corresponding total training steps are $0.1,0.2,0.5$ million)
for InvertedPendulum-v2, Reacher-v2, and Swimmer-v2, respectively. 
The average return over the five evaluation episodes is regarded as the episodic return $\mathcal{R}_n^w$ 
for the training unit $n=1,\dots, N$. For each algorithm, this process is repeated for five times with $w=1,\dots,5$.
Each repeated run is done with a different random seed. The same set of five random seeds are used to compare all the algorithms in a certain environment.

The evolution curves of the episodic return versus the number of training units (in terms of the total number of training steps) are plotted in the figures in this section.
A solid curve indicates the mean of the five repeated runs, i.e., $\bar{\mathcal{R}}_n =  \sum_{w=1}^5 \mathcal{R}_n^w /5$
and the shaded area shows the confidence interval of the repeats representing the corresponding standard devision $\sigma_n$. 
  It is worth mentioning that each curve is smoothed using its moving average of $20$ training units for clarity of understanding. 
The results for the three different environments are discussed in the following subsections.

\begin{table}[h!]
\caption{Design parameters. For common parameters used across different algorithms, same value is used and hence are not repeated.}
\centering
\begin{tabular}{ |l|l|}
\hline
Parameter & Value \\
\hline
\textbf{SAC} & \\
Learning rate & $ 5\times 10^{-4}$ \\
Optimizer & Adam \cite{kingma2014adam} \\
Entropy  regularization  coefficient ($\alpha $) &  adaptive\\
Number of hidden units per layer & $ 256$\\
Experience Replay buffer size  & $10^6$\\
Soft update factor & $10^{-2}$\\
Batch size & $50$ \\
Discount factor ($\gamma$) & $0.99$ \\
Gradient step & $1$ \\
\hline 
\textbf{PER} & \\
$\beta_1$ (as $\alpha$ in \cite{schaul2015prioritized}) & 0.6 \\
$\beta_2$ (as $\beta$ in \cite{schaul2015prioritized}) & 0.4 (initial)\\
\hline
\textbf{ERE} & \\
$\eta_0$ & $0.996$ \\
$\eta_1$ & $1.0$ \\
\hline
\textbf{SDP+MO/O} & \\
SDP threshold value ($\zeta_{th}$)  & $0.5$ \\
Delay length ($\xi$) & $10$\\
\hline
\end{tabular}
\label{table:parameter}
\end{table}

\subsection{InvertedPendulum-v2 }
 
\subsubsection{Environment} The task in this environment is to balance a pole hinged to a cart that sits on a rail and is moved by externally applied force. The four dimensional state space consists of the position and velocity of the cart and the angle and angular velocity of the pole. The one dimensional action space represents the force applied by the cart's actuator.  
The reward is $+1$  per time-step until an episode completes.
An episode is defined as a specified number of steps, e.g., $1,000$ in our experiments, if
the pole does not topple over in these steps; or prematurely terminated when the pole topples over.
So, the maximum possible episodic return is 1,000 in this task.

\begin{figure}[t]
\centering
  \begin{subfigure}{.5\textwidth}
   \centering
  \includegraphics[width=.8\linewidth]{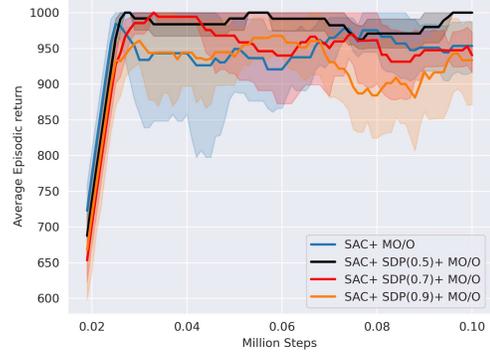} 
   \caption{} 
   \label{SDE_variant_1}
  \end{subfigure}
  \begin{subfigure}{.5\textwidth}
   \centering
  \includegraphics[width=.8\linewidth]{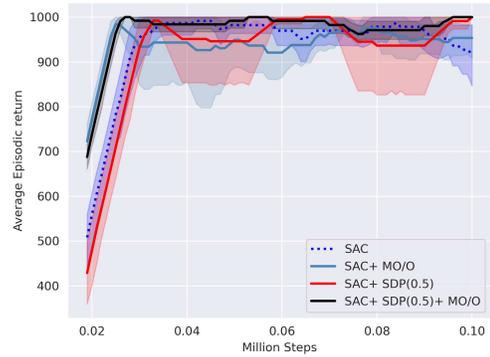} 
   \caption{}
   \label{SDE_variant_2}
   \end{subfigure}
\caption{Evaluation of ISAC artifacts on InvertedPendulum-v2. (a) Performance of ISAC with different SDP threshold $\zeta_{th}$; (b) Performance comparison with different artifacts of ISAC. }
\end{figure}

 \subsubsection{Evaluation of ISAC}
The effect of the SDP threshold value $\zeta_{th}$ on the performance of ISAC is demonstrated in Fig.~\ref{SDE_variant_1}.  We can see that a high threshold value $(\zeta_{th} = 0.9)$  leads to an overfitted/ myopic performance by the policy. After similar hyperparameter exploration, we found that $\zeta_{th} = 0.5$ works best for all the environments. 
The individual contribution of the SDP and MO/O functions of ISAC is presented in Fig. \ref{SDE_variant_2}. 
It is observed that that only running SDP type prioritization on SAC without MO/O based on-policy data mixing (i.e. SAC+ SDP) leads to slow learning of the policy. Whereas adding the MO/O artifact, i.e., SAC+ SDP+MO/O, substantially speeds up the performance. While running ISAC without SDP based prioritization (i.e. SAC+ MO/O), the performance of the policy does not show high performance.

\begin{table*}[t]\centering
\caption{Maximum episodic return and average standard deviation ($\bar{\mathcal{R}}_{max}$, $\bar\sigma$).}
\label{table: performance max values}
\begin{tabular}{|l|l|l|l|}
\hline\hline
\textbf{Environment} & InvertedPendulum-v2 &  Reacher-v2 & Swimmer-v2\\ 
\hline \hline
\textbf{SAC} & $1000, 79.704 $   &  $ -3.790 , 0.962$ & $87.751, \mathbf{11.821}$ \\
\textbf{SAC+ PER} & $1000, 94.213 $   &   $ -3.843 , 0.969$ & $62.990, 13.124$ \\
\textbf{SAC+ PER+ ERE} & $ 1000 , 72.783 $  & $ -3.757 , 0.980$ & $57.626, 13.645$ \\
\textbf{ISAC} & $ 1000, \mathbf{30.578}$   &  $\mathbf{-3.673,  0.859}$ & $\mathbf{108.969}, 19.119$ \\
\hline \hline
\end{tabular}
\end{table*}

\begin{table*}[t]\centering
\caption{Training steps (average and standard deviation of   $\mathcal{T}^w$ for $ w=1,\dots,5$) to reach a prescribed target score.  }
\label{table: Length of training}
\begin{tabular}{|l|l|l|l|}
\hline\hline
\textbf{Environment} & InvertedPendulum-v2 & Reacher-v2 & Swimmer-v2\\ 
\hline \hline
\textbf{Target score} & $954.467$   &  $-4.735$& $ 53.223$ \\
\hline
\textbf{SAC}  & $9,200 \pm 1,720$   &  $ 30,100 \pm 3,878$ & $34,600 \pm 34,719$ \\
    \textbf{SAC+ PER}   & $9,200 \pm 3,249$   & $ \mathbf{26,100 \pm 7,742 } $ & $57,000 \pm 27,224$ \\
\textbf{SAC+ PER+ ERE}  & $ 7,400 \pm 2,416$  &  $ 33,500 \pm  13,311$ & $51,200 \pm 28,491$\\
\textbf{ISAC} & $ \mathbf{6,800 \pm 748} $   &  $ \mathbf{26,500 \pm 5,630} $ & $\mathbf{19,600 \pm 8,616}$ \\
\hline \hline
\end{tabular}
\end{table*}

\begin{figure}[t]
\begin{subfigure}{.5\textwidth}
  \centering
  %
  \includegraphics[width=.8\linewidth]{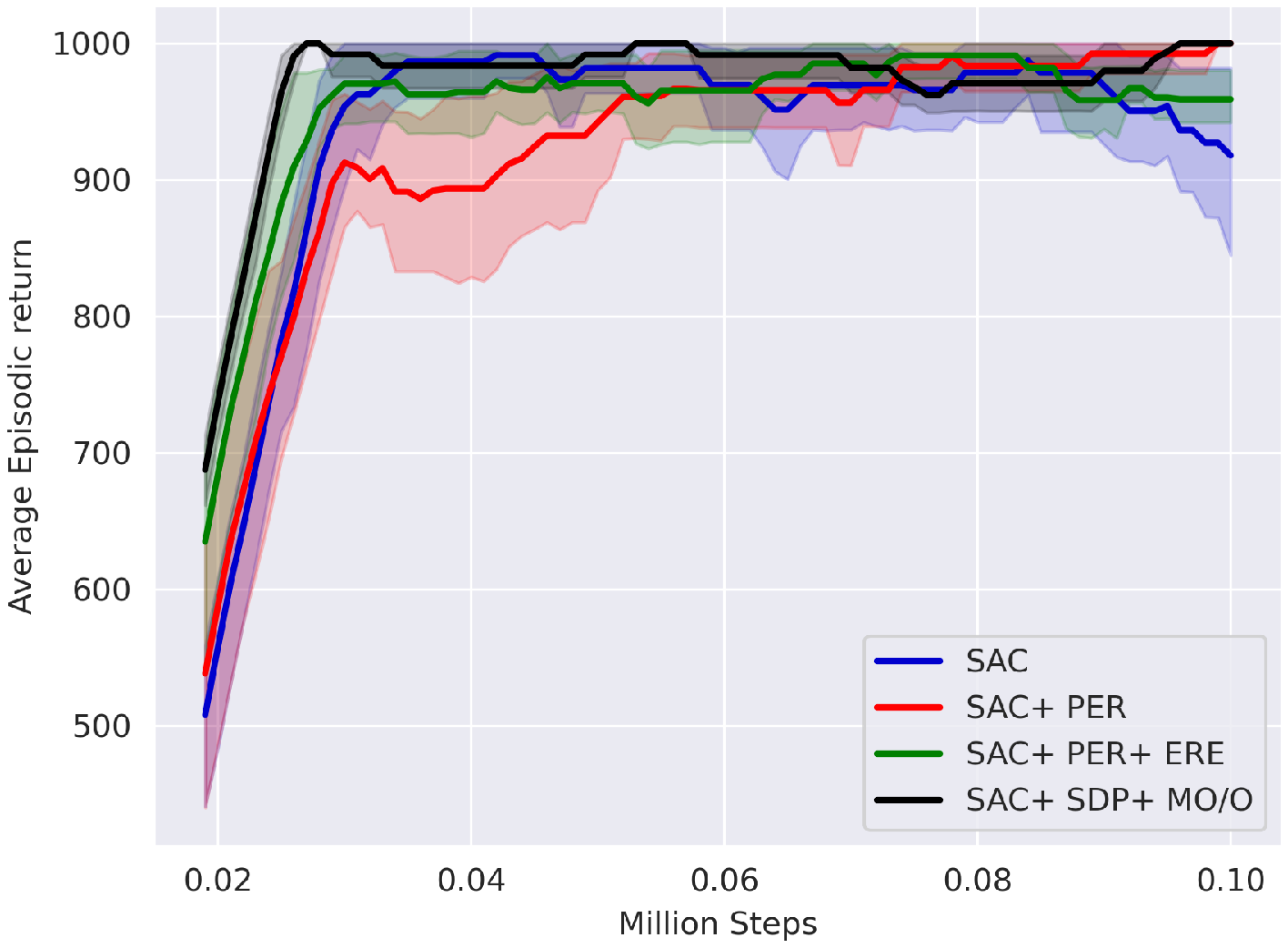}  
  \caption{InvertedPendulum-v2}
  \label{fig:sub-first}
\end{subfigure}
\begin{subfigure}{.5\textwidth}
  \centering
\includegraphics[width=.8\linewidth]{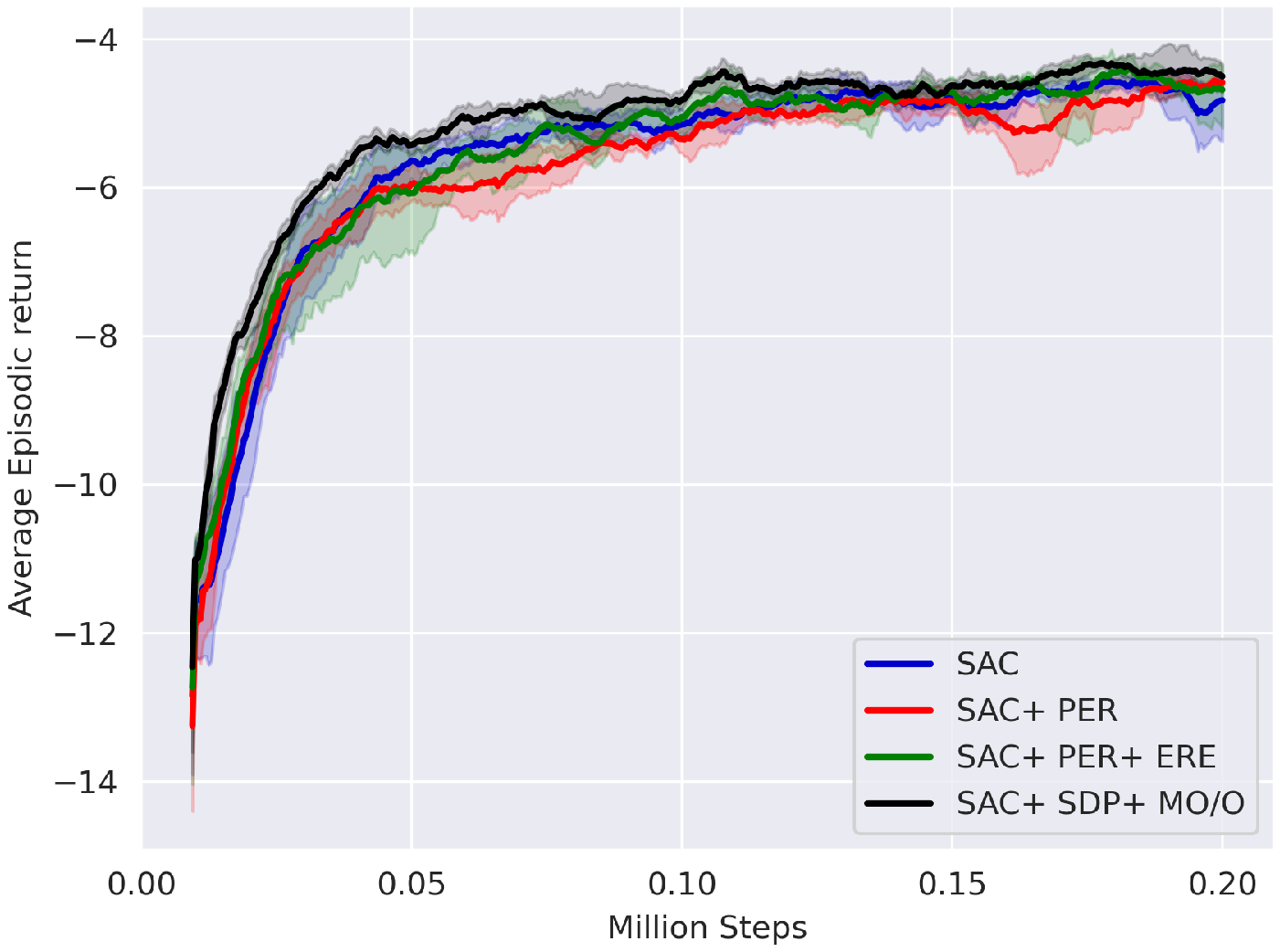} 
  \caption{Reacher-v2}
  \label{fig:sub-second}
\end{subfigure}
\begin{subfigure}{.5\textwidth}
  \centering
\includegraphics[width=.8\linewidth]{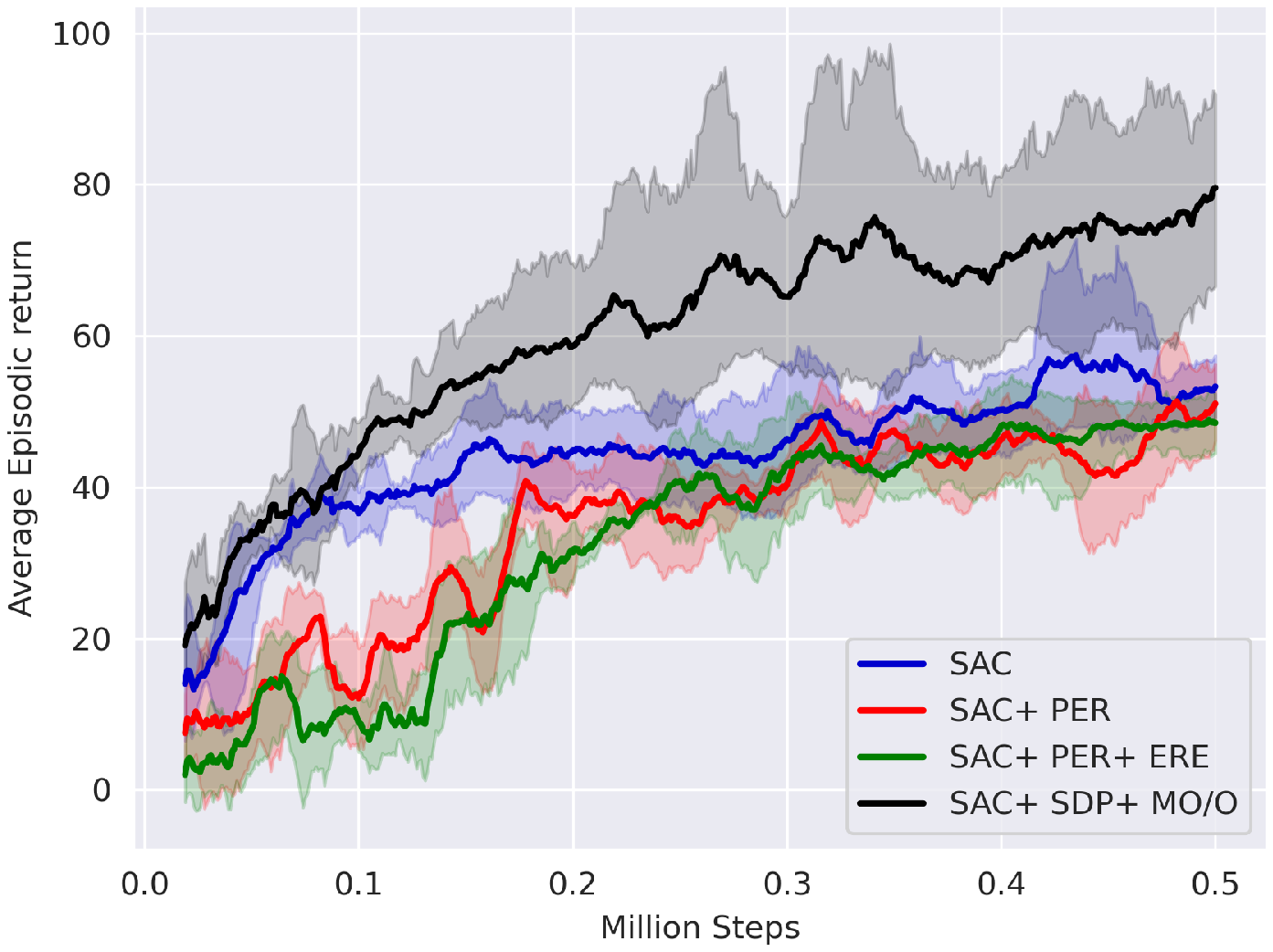}
  \caption{Swimmer-v2}
  \label{fig:sub-third}
\end{subfigure}
\caption{Performance comparison of ISAC with other benchmarks.
}
\label{fig:compare_main}
\end{figure}

\subsubsection{Comparison with benchmarks}

From the performance plots in Fig.\ref{fig:sub-first} it is clear that ISAC significantly outperforms the other three benchmark algorithms. The crucial characteristics of ISAC  is its fast rise to near optimal performance, during the early stage of learning. The performance also includes good stability of the training progresses that keep on improving gradually with minor fluctuations.   The comparison is also quantitatively shown in Tables~\ref{table: performance max values} and \ref{table: Length of training}.
In Table \ref{table: performance max values} we present the maximum attained episodic return 
averaged over the five repeats, i.e., $\bar{\mathcal{R}}_{max} =\max_{n=1}^N \{ \bar{\mathcal{R}}_n \}$,
that demonstrates the ability of an algorithm in achieving the peak performance. In the table, we also present the average of the 
standard deviation of the five repeats for the full training period, i.e., $\bar\sigma = \sum_{n=1}^{N} \sigma_n /N$
that demonstrates the robustness of the performance.

Let $\mathcal{T}^w$ be the number of training steps taken by a certain algorithm to reach a set target score. 
The  target score is set as the average episodic return of the  SAC benchmark over its final $N_f$ units,  i.e.,  $\mathcal{R}_{target} = \Sigma_{n=N-N_f}^N\{ \bar{\mathcal{R}}_n \}/ N_f,$
where we use $N_f=50, 100, 50$ for InvertedPendulum-v2, Reacher-v2, and Swimmer-v2, respectively, 
all corresponding $0.05$ million steps of training.
In Table \ref{table: Length of training} we present the average and standard deviation of  $\mathcal{T}^w$ for the five repeats $w=1,\dots,5$, as a measure of an algorithm's sample efficiency  to reach near-optimal performance. 

As the maximum possible episodic return is 1,000 in this task, there is no difference in the index $\bar{\mathcal{R}}_{max}$.
ISAC is at the top the list of average standard deviation $=30.578$ while the next best value is $72.783$ for SAC+PER+ERE. In Table \ref{table: Length of training} we can see that ISAC takes significantly less training steps to reach a set target score. 
Again SAC+ PER+ ERE scores the second best which took $600$ more steps in average but with more than three times 
standard deviation.

\subsection{Reacher-v2 }
\subsubsection{Environment}  Reacher-v2 represents an environment of a two DOF robotic arm anchored to the centre of a square arena and a randomly generated target. The arm consists of two links of equal length and two actuated joints. The dynamic model is described in an eleven dimensional state space which consists of joint angles (4 dimensions), coordinates of the target (2), end-effector's velocity (2), and coordinates of the vector from the target to the end-effector (3). The two dimensional action space is for  the torques applied to the two actuators. The action value of each actuator is continuous in time within the range $[-1.0, 1.0]$.
The control objective is to make the end-effector of the arm reach the randomly generated target as fast as possible within $50$ steps, i.e., one episode. The reward function consists of two reward components: (i) r-distance: reward based on how close the end-effector is to the target, and (ii) r-action: reward based on the torques used at the actuators to manipulate the arm, that is, 
\begin{align*}
    R(a_t,s_{t+1}) = \underbrace{-\| d_{t+1}\| _{2}^{2}}_{\text{r-distance}}\, \underbrace{-\| a_t\| _2^2}_{\text{r-action}}
\end{align*}
where $d_{t+1}$, the distance between the end-effector and the target, is a function of the state vector $s_{t+1}$, 
and $a_t$ is the action vector.

\subsubsection{Comparison with benchmarks}  
As seen in the comparison plots in Fig. \ref{fig:sub-second} the performance of ISAC is superior than the compared benchmarks. It achieves $\bar{\mathcal{R}}_{max}$ of $-3.673$ while its nearest contender SAC+ PER+ ERE scores $-3.757$ and SAC's performance is further down at $-3.790\pm 0.962$; see Table \ref{table: performance max values}. The average standard deviation 
for ISAC is also smaller than the three benchmarks. 
As shown in Table \ref{table: Length of training}, ISAC and SAC+PER can reach near-optimal performance using less training steps
than the other two algorithms. SAC+PER marginally beats ISAC in average, but with a higher standard deviation.

\subsection{Swimmer-v2}
\subsubsection{Environment} Swimmer-v2 represents a planar robot swimming in a viscous fluid. It is made up of three links 
(head, body and tail) and two actuated joints connecting them. 
The system dynamics can be described in a ten dimensional state space, which consists of 
position and velocity of center of body (4), 
angle and angular velocity of center of body (2),
and angle and angular velocity of two joints (4). 
The two dimensional action space consists of the torques applied on the two actuated joints. 
The objective in this experiment is to stimulate the maximal forward velocity (the positive x-axis) by actuating the two joints,
with the reward function defined as follows
\begin{align*}
    R(a_t,s_{t+1}) &= v^{x}_{t+1} - 0.0001\, \|a_{t}\|^2_2 
\end{align*}    
where $v_{t+1}^{x}$ (an element of $s_{t+1}$) is the forward velocity  and $a_{t}$ the two dimensional action torques.  
The value of each action torque is continuous in time within the range $[-1.0,1.0]$, out of which the value is clipped to its maximum or minimum value.  For every $1,000$ steps, called one episode, the environment is reset and the swimmer starts at a new random initial state. 
There is no premature termination condition applied to an episode.

\subsubsection{Comparison with benchmarks} Similarly, ISAC performs better compared to the benchmarks as shown in 
Fig.~\ref{fig:sub-third}. In particular, it attains the highest $\bar{\mathcal{R}}_{max}$ score of $108.969$ in Table~\ref{table: performance max values}. The second best score of  $87.751$ is achieved by SAC but with a better standard deviation of $11.821$.
From the comparison in Table~\ref{table: Length of training}), the performance of ISAC is outstanding compared with 
the four benchmarks in terms of the training steps to reach a specified target score.
 

\section{Conclusion and Future Work} \label{concl}

We have proposed an improved  SAC algorithm, i.e., ISAC by introducing two major enhancements on the existing SAC framework. ISAC trains conventional SAC networks using prioritized samples from the ER buffer which are also augmented with the latest on-policy sample at each timestep. The episodic return based prioritization scheme (SDP) and augmentation of samples with latest on-policy experiences (MO/O) are simple to implement on the existing SAC framework. 
It does not require any special data structure to implement ISAC, unlike SAC+ PER whose implementation requires a priority tree. 
ISAC attains superior results on a number of test environments compared to other recent variants of SAC. Our experiments 
have shown that, compared to some benchmarks, ISAC attains higher episodic return and reaches near optimal performance using less training steps.  It is interesting to test the new algorithm for more challenging environments of higher dimensional states and action spaces in the future work.  Further improvement of the  prioritization method, e.g., to ensure that the networks are trained from a higher percentage of newer or \say{less seen before data} samples from the ER buffer, is also to be studied.

\bibliographystyle{IEEEtran}
\bibliography{refer}
\end{document}